\documentclass[letterpaper, 10 pt, conference]{ieeeconf}  

\IEEEoverridecommandlockouts                              

\usepackage{graphicx} 
\usepackage{listings}
\usepackage{spverbatim}
\usepackage{fancyvrb}
\usepackage{lipsum} 
\usepackage[utf8]{inputenc}
\usepackage{array}
\usepackage{xcolor}
\usepackage{hyperref}

\usepackage{cite}
\usepackage{amsmath,amssymb,amsfonts}
\usepackage{graphicx}
\usepackage{textcomp}
\usepackage{graphics} 
\usepackage{epsfig} 
\usepackage{mathptmx} 
\usepackage{times} 
\usepackage{mathrsfs}
\usepackage{mathtools}
\usepackage{stfloats}
\usepackage{algorithm,algpseudocode}
\usepackage{multirow}
\usepackage{subcaption}
\usepackage{url}
\usepackage[thinlines]{easytable}
\usepackage[utf8]{inputenc}
\newcommand{\etal}{\textit{et al.}}
\usepackage{caption}

\usepackage{etoolbox}
\makeatletter
\newcommand*{\algrule}[1][\algorithmicindent]{%
  \makebox[#1][l]{%
    \hspace*{.2em}
    \vrule height .75\baselineskip depth .25\baselineskip
  }
}

\newcount\ALG@printindent@tempcnta
\def\ALG@printindent{%
    \ifnum \theALG@nested>0
    \ifx\ALG@text\ALG@x@notext
    \else
    \unskip
    \ALG@printindent@tempcnta=1
    \loop
    \algrule[\csname ALG@ind@\the\ALG@printindent@tempcnta\endcsname]%
    \advance \ALG@printindent@tempcnta 1
    \ifnum \ALG@printindent@tempcnta<\numexpr\theALG@nested+1\relax
    \repeat
    \fi
    \fi
}
\patchcmd{\ALG@doentity}{\noindent\hskip\ALG@tlm}{\ALG@printindent}{}{\errmessage{failed to patch}}
\patchcmd{\ALG@doentity}{\item[]\nointerlineskip}{}{}{} 
\makeatother

\title{\LARGE \bf
 A Predictive Application Offloading Algorithm Using Small Datasets for Cloud Robotics}

\author{Manoj Penmetcha, Shyam Sundar Kannan, and Byung-Cheol Min
\thanks{The authors are with the SMART Lab, Department of Computer and Information Technology, Purdue University, West Lafayette, IN 47907, USA
	{\tt\small mpenmetc@purdue.edu | kannan9@purdue.edu | minb@purdue.edu}}%
}
\begin{document}
\maketitle
\thispagestyle{empty}
\pagestyle{empty}

\begin{abstract}

Many robotic applications that are critical for robot performance require immediate feedback, hence execution time is a critical concern. Furthermore, it is common that robots come with a fixed quantity of hardware resources; if an application requires more computational resources than the robot can accommodate, its onboard execution might be extended to a degree that degrades the robot’s performance. Cloud computing, on the other hand, features on-demand computational resources; by enabling robots to leverage those resources, application execution time can be reduced. The key to enabling robot use of cloud computing is designing an efficient offloading algorithm that makes optimum use of the robot’s onboard capabilities and also forms a quick consensus on when to offload without any prior knowledge or information about the application. In this paper, we propose a predictive algorithm to anticipate the time needed to execute an application for a given application data input size with the help of a small number of previous observations. To validate the algorithm, we train it on the previous $N$ observations, which include independent (input data size) and dependent (execution time) variables. To understand how algorithm performance varies in terms of prediction accuracy and error, we tested various $N$ values using linear regression and a mobile robot path planning application. From our experiments and analysis, we determined the algorithm to have acceptable error and prediction accuracy when $N>40$.

\end{abstract}  


\section{Introduction}
Robots have made inroads into a variety of complicated application spaces, which developments have mainly been driven by the introduction of highly-sophisticated robots in areas such as industrial robots, personal robots, navigation, and robotic surgery \cite{siau2018building,doi:10.1177/0278364918770733}. Among the major impetuses for the sophistication of such high-level robots are recent advancements in information technology \cite{Wang2019ArtificialIM}. The rapid evolution of technology is led by AI, cloud computing, IoT, blockchain, and other developments, and these technologies have opened up many robotic application spaces that were previously unimaginable \cite{Wang2019ArtificialIM}. 

One thing that is common to all these new technologies is that they require high computational resources for application execution. Meanwhile, robots themselves often come with fixed computing capabilities and so cannot fulfill the computational demands of these systems and other new technological advancements. Providing robots with access to external computational resources such as cloud computing is an effective means of solving this problem \cite{article151}. Cloud computing can provide access to on-demand computational resources and significantly enhance application performance for robots that otherwise have only limited computational resources. 

\begin{figure}[t]
    \centering
    \includegraphics[width=\linewidth]{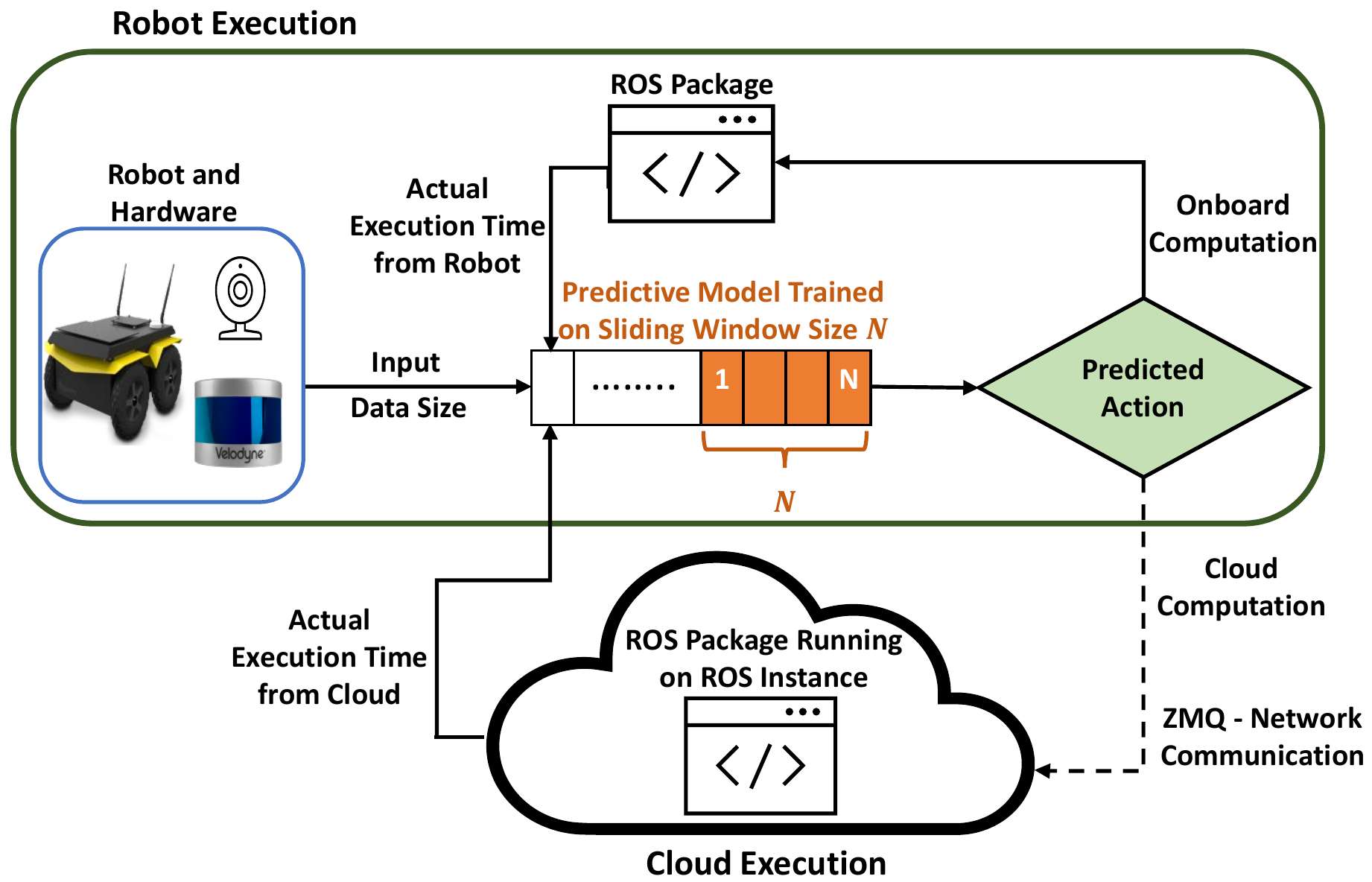}
    \caption{Illustration of the proposed predictive algorithm for offloading decision-making. (Top) Input data size from robot hardware (e.g. LiDAR, camera) along with actual execution times from cloud and robot executions are used to train the predictive model. The training set consists of the previous $N$ values to facilitate learning on the fly and faster training. (Bottom) An instance of ROS running on the cloud subscribes to ROS topics from the robot and responds with application output.}
    \label{fig:genericframework}
    \vspace{-1.2em}
\end{figure}

James J. Kuffner in 2010 \cite{10031099795} was the first to introduce the term \textit{Cloud Robotics}, and simultaneously explained the advantages of enabling cloud resource access for robot platforms. Notably, although robot onboard resources are often fixed, modern robots do typically come with a fair amount of computational resources, and it is important to consider these resources in offloading decision-making. Ideally, the offloading algorithm should make a decision based on cost parameters such as execution time, energy usage, and CPU availability. Researchers have recently proposed several dynamic offloading solutions that use machine learning algorithms and consider cost parameters \cite{8403209}. However, these machine learning algorithms require training on large datasets to make accurate predictions. Also, no two robots and applications are the same, thus a machine learning model trained to work for a certain type of application and robot is not guaranteed to work well for other combinations, and such models are not easy to generalize. On top of all that, obtaining large datasets is itself a challenging task. Hence, it is important to also consider an algorithm that can be trained quickly and learn on the fly using smaller datasets. We introduce here a predictive algorithm, outlined in Fig. \ref{fig:genericframework}, that is designed to help robots make offloading decisions without any prior knowledge about the application.

The main contributions of the paper are as follows: 
\begin{itemize}
\item We introduce a predictive algorithm to predict the execution time of an application under both cloud and onboard computation, based on the application input data size. We further validate the predictive algorithm with a linear regression model.
\item The algorithm is designed to be trained after the application has been initiated (online learning). To make the algorithm faster to train, we train it on a fixed dataset of $N$ previous observations. We experiment with various $N$ to observe the prediction accuracy and error. 
\item We employ Gazebo simulation to analyze the proposed algorithm using a robot path planning application. 
\end{itemize}

The remainder of the paper is organized as follows: In Section \ref{sec:literature}, we describe related work on application offloading algorithms. In Section \ref{sec:generalized}, we introduce a predictive algorithm. In Section \ref{sec:linearregression}, we present linear regression and a mobile robot path planning application to validate the generalized algorithm. In Section \ref{sec:results}, we provide extensive experimental results and analysis. Finally, we conclude our work and present our future directions in Section \ref{sec:conclusion}. 

\section{Literature Review} \label{sec:literature}
For robots with bare minimum computational capabilities, full application offloading will be an obvious choice. However, many robots currently produced are computationally capable, hence are capable of onboard execution of many applications; this leads to dynamic application offloading being more appropriate, with offloading decisions taking into account both the robot’s computational capability and the application’s computational requirements. Hence, application offloading approaches can be broadly categorized into full offloading and dynamic offloading. 

Since the inception of cloud robotics in 2009, several architectures have been proposed that mainly focused on full offloading; these cater to specific applications such as navigation, image classification, and localization \cite{7006734, 7482658, 9283148, 6853392}. 

To our knowledge, the area of dynamic application offloading for cloud robotics is understudied, with only a few studies published to date that focus on dynamic application offloading. In one example, Rahman \etal~proposed an offloading solution for cloud-networked multi-robot systems that was based on a genetic algorithm and focused on energy efficiency as the criterion informing decision-making \cite{ RAHMAN201911}. Alli \etal~proposed an offloading solution based on the Neuro-Fuzzy model, a machine learning method, which aims to minimize latency and energy for smart city applications \cite{ ALLI2019100070}. Some other solutions based on edge computing and machine learning have also been proposed \cite{ 9291664,9113305}.  

In the last couple of years, additional approaches have tried applying deep reinforcement learning (DRL) \cite{ lillicrap2019continuous} -based algorithms for making application offloading decisions. Chicachali \etal \cite { DBLP:journals/corr/abs-1902-05703} used such a strategy to offload an object detection application to the cloud, while Peng \etal \cite {peng2020multi} used a DRL-based Deep-Q-Network algorithm for offloading based on energy consumption and task makespan.

In reviewing the literature, we observed various extant solutions based on genetic algorithms, machine learning, and DRL. All of these algorithms require large datasets and substantial time in training to achieve convergence and make accurate predictions. Also, there is no guarantee that the trained algorithm will work for other types of applications and robots. Hence, it is important to design a lightweight offloading algorithm that can learn after the application is initiated. Our proposed algorithm is mainly designed to being adaptable to diverse applications and not requiring any pretraining. As the proposed algorithm is not pre-trained, it is important for it to quickly learn optimal decision-making. To make the algorithm faster to train, we train it on a fixed dataset of $N$ previous observations. We experiment with various $N$ to observe the correlation, prediction accuracy and error.


\section{Generalized Predictive algorithm} \label{sec:generalized}

In this section, we introduce a generalized version of the predictive algorithm ($\Psi$) for online application offloading. The simple nature of the proposed predictive algorithm empowers the user to proceed without any of the pre-computing that is usually required for machine learning algorithms or other optimizing techniques. This paper is interested in a useful, generalized dynamic offloading algorithm that can be applied to robotic applications and has an independent variable that can used to predict a dependent variable. 

An application execution usually consists of subtasks, where each task can be independent or dependent on another. These tasks process the received input and give a resulting output. Here, we categorize an application $A$ into application tasks $a_i$, where $A \in a_i $ and $i$ is the application task sequence number, and for simplicity we assume that application tasks are executed in sequence. The time needed for execution of an application depends on its input arguments, such as the size of the data ($d_i$). Our basic premise is that we can predict both cloud execution ($p_i^{c}$) and local execution ($p_i^{l}$) times using the input arguments provided to the application, thus all applications can be generalized as the sum of all components dependent on application arguments. These predicted values $p_i^{c}$ and $p_i^{l}$ can be used in offloading decision-making, and generally, we can represent the offloading choice as: 
\begin{align}
    & \text{Cloud computation} \quad\quad when \quad p_i^{c} < p_i^{l} \nonumber \\
   & \text{Onboard computation}   \quad when \quad p_i^{c} \geq p_i^{l}. \label{localexecutiontimere}
\end{align}

To compare $p_i^{c}$ and $p_i^{l}$, a predictive function is needed; however, estimating execution time can be simple or complex. Many algorithms and functions can be created to predict execution time by means of an algorithm built from the values of the application’s input variables. For small systems with straightforward relationships, $n$-dimensional regressions are the obvious choice; meanwhile, for systems with many variables and complex relationships, machine learning is commonly used. Some of the most widely-used predictive algorithms are linear regression \cite{montgomery2021introduction}, random forest \cite{doi:10.1021/ci034160g}, time-series algorithms \cite{wei2006time}, and k-means clustering \cite{LIKAS2003451}. The time complexity of an application with one major variable can be written as $O(u)$, where $u=1$, $n$, $n$log$n$, $n^2$, $2^n$, etc. For an application carrying out the same task repeatedly on a dataset of constant size, the complexity is $O(1)$. For an application using only one variable, regardless of complexity, estimating $p_i^{c}$ and $p_i^{l}$ with regression from sample data coordinates is straightforward. 

One key aspect of the proposed predictive algorithm is that it includes a fixed number $N$, the window size of the most recent execution data (cloud and local) that is used for training the algorithm to form a consensus and predict the execution times $p_i^{c}$ and $p_i^{l}$. Algorithm \ref{regressionalgo} describes the predictive model, where we use First-In First-Out (FIFO) queue data structures $B^{l}$ and $B^{c}$ to store the $N$ training values. Having a fixed training size keeps memory low and keeps prediction runtimes low. By retaining the most recent observations, the algorithm can also account for exogenous factors that influence execution time. For example, even though we define that application execution time depends on the input arguments, there will be instances where the execution time of the application varies based on external factors not captured in the input arguments, such as network loss and high CPU usage by other applications. By storing the $N$ most recent executions, we can capture some of the external factors that affect execution time. After training the algorithm with the stored queue values, the algorithm will receive current state values and predict execution times that can be used for offloading decision-making. 

	\begin{algorithm}
		\caption{Predictive Algorithm for application offloading}
		\label{regressionalgo}
		\begin{algorithmic}[]
			\State Initialize FIFO based Queues $B^{l}$ and $B^{c}$ with size $N$;
            \For{For application task sequence $i$ = 1}
                    \Repeat 
                        \State Fit $\psi^{l}$ and $\psi^{c}$ with $B^{l}$ and $B^{c}$;
                        \State Load input data size $d_i$ for $a_i$;
                        \State Predict $\psi^{l}(d_i) = p_i^{l}$ and $\psi^{c}(d_i) = p_i^{c}$;
                        \If{$size(B^{l}) \text{ and } size(B^{c}) == N$}
                            \If{$p_i^{l} < p_i^{c}$}
                            \State Execute the application onboard;
                            \State Calculate $t_i^{l}$;
                            \State Append $B^{l}$ with $(t_i^{l}, d_i)$;
                            \Else
                            \State Execute the application on AWS;
                            \State Calculate $t_i^{c}$;
                            \State Append $B^{c}$ with $(t_i^{c}, d_i)$; 
                            \EndIf
                        \Else
                            \State Execute application onboard and AWS;
                             \State Calculate $t_i^{l}$ and $t_i^{c}$;
                             \State Append $B^{l}$ with $(t_i^{l}, d_i)$;
                             \State Append $B^{c}$ with $(t_i^{c}, d_i)$;
                        \EndIf
                        \State Set $i = i+1$;
                    \Until{{Application is terminated;}}
            \EndFor
		\end{algorithmic}
	\end{algorithm}
	
\section{Linear Regression and Mobile Robot Path Planning Application} \label{sec:linearregression}

In this section, we introduce the linear regression and the mobile robot path planning application that we use to validate the proposed online learning algorithm. 

\subsection{Linear Regression Model}

The goal of the algorithm (Algorithm \ref{regressionalgo}) is to predict $p_i^{c}$ and $p_i^{l}$ (dependent variables) from an independent variable that has some correlation with execution time. Application execution time is proportional to the input data size, and there exists a linear relation between them \cite{8377343}. Hence, we want to predict execution time from the input data size ($d_i$). For predictions involving linear relationships, $n$-dimensional regressions are the obvious choice, hence we used a lightweight linear regression-based predictive model ($\psi$) to predict $p_i^{c}$ and $p_i^{l}$ from $d_i$. We term this algorithm lightweight is because we train it on a fixed dataset of size $N$. The duration of training will thus be considerably shorter relative to other models trained with larger datasets. Finally, to validate the proposed algorithm, we need a dataset with actual application execution times for both cloud computation ($t_i^{c}$) and local computation ($t_i^{l}$) and also the corresponding input data sizes ($d_i$). We derived this information for a mobile robot path planning application, the process of which will be explained in sections \ref{sec:Robot Platform} and \ref{sec:cloud Platform}. 

To predict local execution time ($p_i^{l}$), we used linear regression with the least squared method, as follows: 
\begin{equation}
\label{localexecutiontime}
\begin{split}
 p_i^{l} = m^{l} * d_i + c^{l}
\end{split}
\end{equation} 
where $m^{l}$ is the slope and $c^{l}$ is the intercept, which are derived as follows: 
\begin{equation}
\begin{split} 
 & m^{l} = \frac{\sum_{i=1}^{N} (d_i -\overline{d}) * (t_i^{l} - \overline{t^{l}})} {\sum_{i=1}^{N} (d_i -\overline{d})^{2}}\\[3pt]
 & c^{l} = \overline{t^{l}} - m^{l} * \overline{d},
\end{split}
\end{equation} 
where $N$ is the size of the window of previous observations used for fitting the linear regression model, $\overline{d}$ is the mean of the input data size, and $\overline{t^{l}}$ is the  mean of the actual execution time from $N$ previous observations. 

Similarly, to predict cloud execution time ($p_i^{c}$), we can represent the linear regression equation with the least squared method as follows: 
\begin{equation}
\label{cloudexecutiontime}
\begin{split}
 p_i^{c} = m^{c} * d_i + c^{c}
\end{split}
\end{equation} 
where $m^{c}$ is the slope and $c^{c}$ is the intercept, which are derived as follows: 
 \begin{equation}
\begin{split} 
 & m^{c} = \frac{\sum_{i=1}^{N} (d_i -\overline{d}) * (t_i^{c} - \overline{t^{c}})} {\sum_{i=1}^{N} (d_i -\overline{d})^{2}}\\[3pt]
 & c^{c} = \overline{t^{c}} - m^{c} * \overline{d},
\end{split}
 \end{equation} 
where $\overline{t^{c}}$ is the  mean of the actual execution time from $N$ previous observations. 

\begin{figure*}
  \centering
  \includegraphics[width=\textwidth]{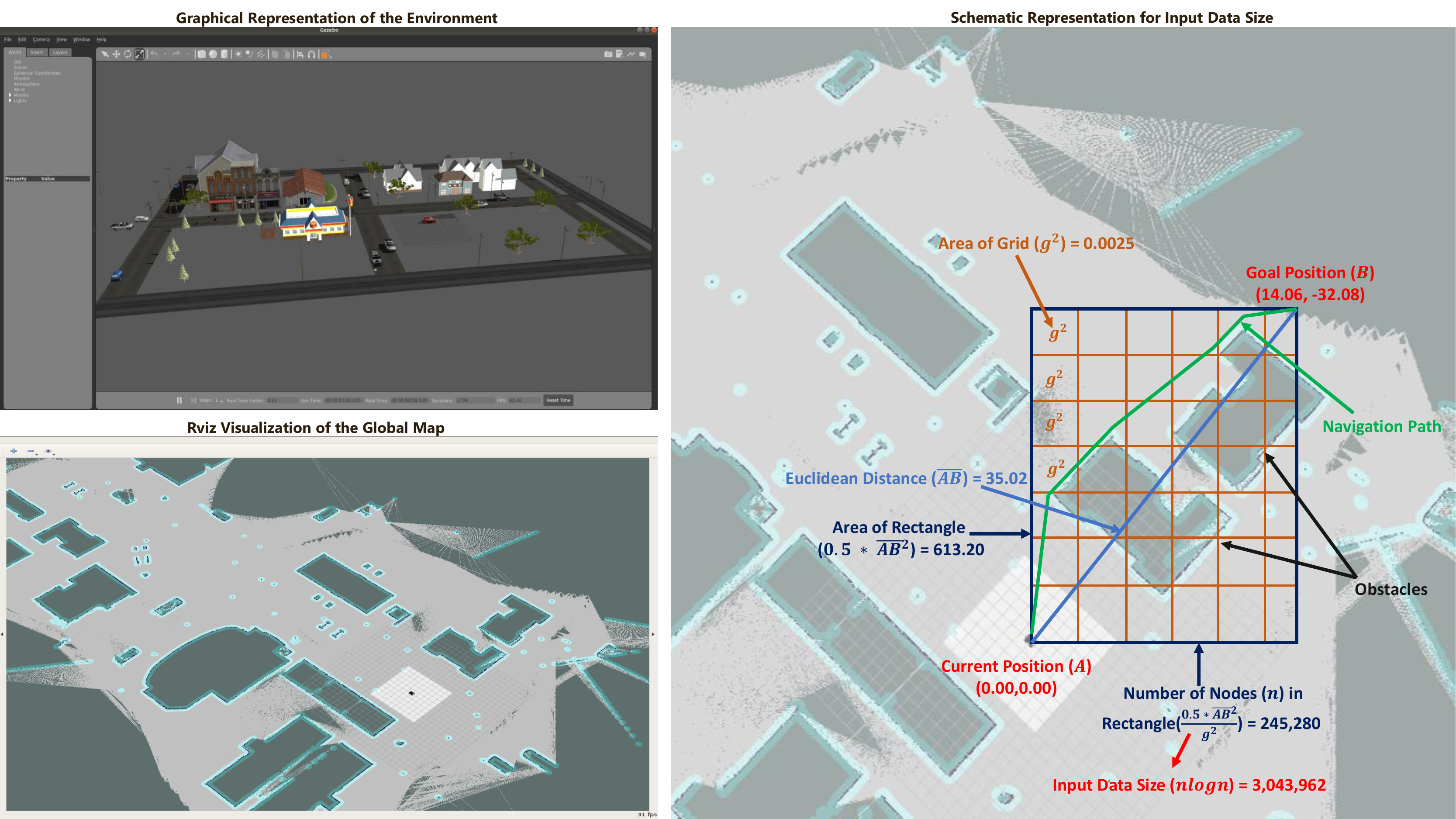}
  \caption{The left panel shows the graphical simulation and global map of the environment that was used to validate the proposed algorithm. The right panel illustrates a representative scenario for calculating the number of nodes that the robot (a Husky from Clearpath Robotics) needs to explore before reaching its destination. In this instance, the starting position $A$ is (0.00, 0.00) and the goal position $B$ is (14.06, -32.08). The optimal path between those points is represented by a green line, and the Euclidean distance is $\overline{AB}$ = 35.02. The area of the rectangle having $\overline{AB}$ as diagonal is 613, and the number of nodes that rectangle contains ($n$), per Eq. \ref{planningsize1}, is 245,280. Thus, the input data size ($d$ = $n$log$n$) is 3,043,962.}
  \label{fig:gazebo}
  \vspace{-0.2em}
\end{figure*}

\subsection{Robot path planning platform}
\label{sec:Robot Platform}
\begin{figure}[t]
    \centering
    \includegraphics[width=\linewidth]{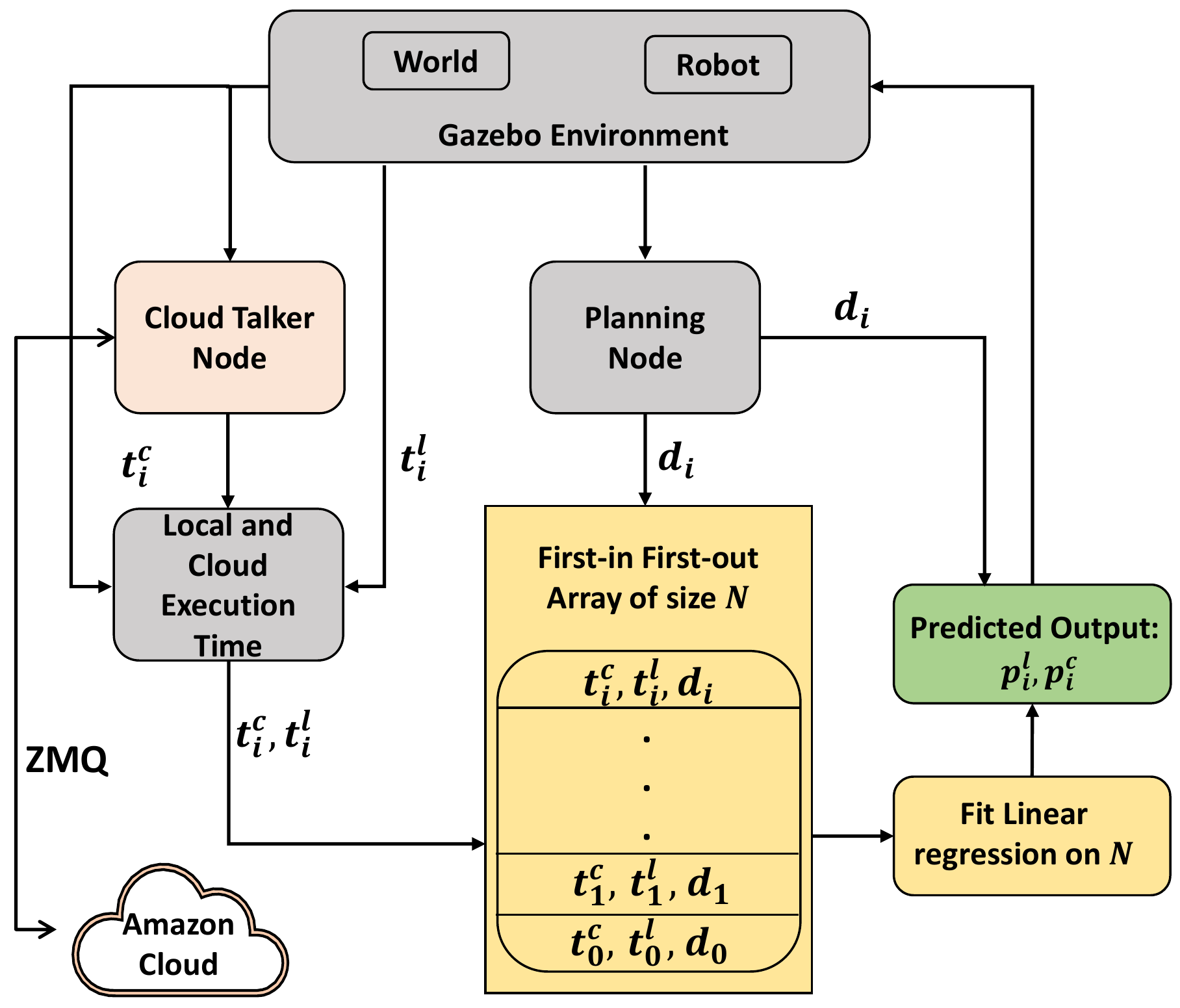}
    \caption{Navigation framework for algorithm validation. A mobile robot performs path planning inside a Gazebo world by interacting with AWS. The data from the simulation are fitted to a regression model to predict execution times for a given input data size.}
    \label{fig:navigationframework}
    \vspace{-1em}
\end{figure}

To validate the proposed algorithm, we used a Gazebo simulation environment, as depicted in Fig. \ref{fig:gazebo}. By using the proposed solution, robots can offload a navigation application to the cloud depending on the execution time required. To validate the proposed algorithm in a simple and practical environment, we considered a basic path planning application. Using the framework shown in Fig. \ref{fig:navigationframework}, we obtain actual execution times $t_i^{c}$ and $t_i^{l}$ for the corresponding $d_i$ for a robot path planning application. Based on the predicted execution times $p_i^{c}$ and $p_i^{l}$, the robot executes the application either on the cloud or locally. Afterwards, the robot stores actual execution times ($t_i^{c}$ and $ t_i^{l}$) along with corresponding data size ($d_i$) values in a queue of size $N$. Finally, the predictive algorithm $\Psi$ iteratively updates with new values from that queue for each prediction. 



Here, we discuss how we derived the application input data size ($d_i$). For path planning, the robot needs to have a map of its environment and also needs to be capable of performing simultaneous localization and mapping (SLAM) in the given environment to obtain an obstacle-free path from the origin to the destination \cite{GALCERAN20131258}. In our experiment, we used Dijkstra’s algorithm \cite{10.1007/BF01386390} to compute the shortest path. When the Robot Operating System (ROS) launches a navigation module, the granularity of the occupancy grid ($g$) is usually set to 0.05 meters \cite{ROSNotes96:online}, but can be manually changed as required. The number of grids that the robot needs to traverse to find a path from the origin ($A$) to the destination ($B$) can be represented as $n$, where $n$ can be roughly represented as the area of a square with diagonal $\overline{AB}$ divided by the area of the grid ($g^{2}$). The mathematical representation of $n$ is as follows: 

\begin{equation}
\label{planningsize1}
    n = \frac{(0.5 * \overline{AB}  ^{2})} {g^{2}}.
\end{equation}

The robot traverses the grids $n$ for several times to find the shortest path, hence the input data size can be represented as $n$log$n$ \cite{goldberg1996expected}, meaning that a reasonable approximation of input data size $d_i$ for robot path planning is:

\begin{equation}
\label{planningsize2}
    d_i = nlogn.
\end{equation}

Fig. \ref{fig:gazebo} demonstrates how we calculated $n$. The starting position of the robot was (0.00, 0.00) and the goal position was (14.06, -32.08), for which the Euclidean distance $\overline{AB}$ can be calculated as 35.02. The area of the rectangle with $\overline{AB}$ as its diagonal can be calculated as 613.20 ($0.5* \overline{AB} ^{2}$). Under the default grid size, each grid block ($g^{2}$) has area 0.0025. Thus, by using the Eq. \ref{planningsize1}, we get the value of $n$ as 245,280, and using Eq. \ref{planningsize2} the input data size $d_i$ as 3,043,962. In the experiment, we normalized the value of $d_i$ by dividing by the global map size, which yielded values in the range of 0 to 5.   

\subsection{Cloud Platform}
\label{sec:cloud Platform}

On the cloud side of the framework, we used Amazon Web Services (AWS) with an Ohio instance. The simulation was carried out on a laptop with a stable wired internet connection having speed greater than 400 Mbps. The average latency observed for data making a round trip between cloud and robot was around 30 milliseconds. To establish communication between the robot and AWS, we used the ZMQ communication protocol \cite{ ZeroMQ83:online}, which is a high-performance asynchronous messaging library that provides a ROS-like publisher-subscriber functionality. These along with other functionalities \cite{Introduc99:online} made ZMQ an ideal choice for our application. 

On AWS, we had a ROS instance running that subscribed to ROS topics from the robot. In this experiment, the robot published LiDAR data and its current and goal positions. The ROS instance on the cloud subscribed to these topics and published back a planned path. In parallel, the robot also computed a planned path on its local ROS instance. We calculated the actual execution times $t_i^{c}$ and $t_i^{l}$ by subtracting the time of data publication from the time at which each instance’s planned path was received.

\section{Results} \label{sec:results}

To evaluate the performance of the decision-making algorithm with training queue of varying size $N$, we programmed a linear regression model based on the logic illustrated in Algorithm. \ref{regressionalgo} on top of the navigation framework described in Fig. \ref{fig:navigationframework}. To maintain consistency across the results, we used a singular dataset that consisted of 1,000 rows with actual execution times $t_i^{l}$, $t_i^{c}$ and corresponding $d_i$, and evaluated training with $N= 5, \,10, \,20, \,30, \,40, \,50, \,75, \,100,$ and $500$ samples. We did not consider queue of size $N <5$, as with so little data there is a good chance of underfitting and insufficient variation for the model to learn. The hardware configuration of the robot and the AWS instance (p2.xlarge) used to generate this dataset are given in Table \ref{tab:hardwareconfig}. 


\renewcommand*{\arraystretch}{1.2}
\begin{table}[htb]
\caption{Hardware configuration of the robot and AWS instance used to generate performance evaluation data.}
\label{tab:hardwareconfig}
\resizebox{\columnwidth}{!}{%
\begin{tabular}{|c|c|c|}
\hline
                   & \textbf{Robot - Local} & \textbf{AWS (p2.xlarge) - Cloud} \\ \hline
\textbf{CPU} &
  \begin{tabular}[c]{@{}c@{}}Intel Core i7-6700 CPU \\ @ 3.40GHz\end{tabular} &
  \begin{tabular}[c]{@{}c@{}}2.7 GHz (turbo) Intel Xeon \\ E5-2686 v4\end{tabular} \\ \hline
\textbf{GPU} &
  \begin{tabular}[c]{@{}c@{}}1 GeForce GTX 1050 - \\ 768 processing cores and \\ 4 gb of GPU memory\end{tabular} &
  \begin{tabular}[c]{@{}c@{}}1 NVIDIA K80 -\\ 2496 parallel processing cores \\ and 12 gb of GPU memory\end{tabular} \\ \hline
\textbf{RAM} & 16 gb                    & 61    gb                           \\ \hline
\textbf{Cores}     & 8                      & 4                                \\ \hline
\textbf{OS}        & Ubuntu-18.04           & Ubuntu-18.04                     \\ \hline
\end{tabular}%
}
\end{table}

\renewcommand*{\arraystretch}{1.5}	
\begin{table*}[]
\caption{Average correlation, mean, residual, and accuracy results for various $N$ sizes.}
\label{tab:results}
\resizebox{\textwidth}{!}{%
\begin{tabular}{|c|c|c|c|c|c|c|c|c|c|}
\hline
\textbf{$N$} & \textbf{\begin{tabular}[c]{@{}c@{}}Average Correlation\\ $t^{c}$ and $d$ (Pearson $r$)\end{tabular}} & \textbf{\begin{tabular}[c]{@{}c@{}}Average Correlation \\ $t^{l}$ and $d$ (Pearson $r$)\end{tabular}} & \textbf{\begin{tabular}[c]{@{}c@{}}Mean\\ $t^{c}$ (Sec)\end{tabular}} & \textbf{\begin{tabular}[c]{@{}c@{}}Mean\\ $p^{c}$ (Sec)\end{tabular}} & \textbf{\begin{tabular}[c]{@{}c@{}}Cloud \\ Residual (Sec)\end{tabular}} & \textbf{\begin{tabular}[c]{@{}c@{}}Mean\\ $t^{l}$ (Sec)\end{tabular}} & \textbf{\begin{tabular}[c]{@{}c@{}}Mean\\ $p^{l}$ (Sec)\end{tabular}} & \textbf{\begin{tabular}[c]{@{}c@{}}Local \\ Residual (Sec)\end{tabular}} & \textbf{\begin{tabular}[c]{@{}c@{}}Prediction \\ Accuracy (\%)\end{tabular}} \\ \hline
\textbf{5} & 0.31726 & 0.73545 & 0.14016 & 0.17697 & 0.03681 & 0.17527 & 0.19099 & 0.01572 & 60.02 \\ \hline
\textbf{10} & 0.31717 & 0.71970 & 0.14029 & 0.15990 & 0.01961 & 0.17537 & 0.17888 & 0.00350 & 64.95 \\ \hline
\textbf{20} & 0.30455 & 0.71198 & 0.14005 & 0.14595 & 0.00590 & 0.17546 & 0.17807 & 0.00260 & 70.60 \\ \hline
\textbf{30} & 0.30245 & 0.70822 & 0.14005 & 0.14578 & 0.00573 & 0.17546 & 0.18259 & 0.00695 & 73.38 \\ \hline
\textbf{40} & 0.30282 & 0.70922 & 0.14016 & 0.14056 & 0.00039 & 0.17552 & 0.18078 & 0.00525 & 76.32 \\ \hline
\textbf{50} & 0.30273 & 0.70641 & 0.14025 & 0.14061 & 0.00036 & 0.17554 & 0.18027 & 0.00473 & 78.80 \\ \hline
\textbf{75} & 0.29914 & 0.69775 & 0.13991 & 0.139972 & 0.00006 & 0.17519 & 0.177697 & 0.00250 & 79.74 \\ \hline
\textbf{100} & 0.28395 & 0.70290 & 0.14048 & 0.13925 & -0.00123 & 0.17564 & 0.175838 & 0.00019 & 79.62 \\ \hline
\textbf{500} & 0.28060 & 0.69707 & 0.13481 & 0.14012 & 0.00531 & 0.17113 & 0.17649 & 0.00536 & 81.90 \\ \hline
\textbf{\begin{tabular}[c]{@{}c@{}}1,000\\ (Full dataset)\end{tabular}} & \textbf{0.27761} & \textbf{0.69221} & \textbf{0.14011} & \textbf{0.14796} & \textbf{0.00785} & \textbf{0.17530} & \textbf{0.18151} & \textbf{0.00621} & \textbf{83.32} \\ \hline
\end{tabular}%
}
\vspace{-1.5em}
\end{table*}

After initialization of the application, the first $N$ elements are solely used to train the linear regression algorithms ($\psi^{l},\psi^{c}$) that are in turn used to predict execution times ($p_i^{l}$ and $p_i^{c}$) for informing offloading decision making. During execution, the actual time of execution is computed ($t_i^{l}$ or $t_i^{c}$) and, along with the corresponding $d_i$, are appended to the corresponding queue, from which the first (oldest) value is then deleted. The corresponding predictive algorithm $\psi^{l}$ or $\psi^{c}$ will then be refitted based on the updated queue to predict the execution time for the next application task ($a_{i+1}$) from a given input data size. The results of the performance evaluation are presented in Table \ref{tab:results}, and in the next subsections we analyze those results in terms of correlation, residuals, and accuracy.

\subsection{Correlation}

We used bivariate Pearson correlation \cite{freedman2007statistics} to estimate the correlation coefficient $r$, which helps us determine the strength of association between independent ($d$) and dependent variables ($t^{l}$ and $t^{c}$). This analysis suggested a low positive correlation of $r$(1,000) = .27761 between cloud execution time ($t^{c}$) and application input data size ($d$) and a strong positive correlation of $r$(1,000) = .69221 between local execution time ($t^{l}$) and application input data size ($d$). Both coefficient values were statistically significant with $p<0.005$. These correlations imply that with an increase in input data size ($d$), there is also an increase in actual execution time, whether local or cloud ($t^{l}$ and $t^{c}$). Furthermore, the evident stronger correlation of local execution time ($t^{l}$) with input data size ($d$) reflects the limited availability of computational resources in onboard execution; when application input data size increases, the application required more computational resources than the robot could accommodate, hence extending execution time. In contrast, cloud execution is better supported in terms of computational resources, and an increase in application input data size ($d$) resulted in a smaller increase of the cloud execution time ($t^{c}$). 

We additionally investigated the impact of $N$ on the correlation between dependent and independent variables. Ideally, as $N$ increases, correlation values should remain the same or become stronger, reflecting better prediction values.  We calculated correlation values for a moving window size of $N$ across the training dataset, then averaged the results to obtain an average correlation value for the given $N$. As seen in column 2 and 3 of Table \ref{tab:results}, the average correlation values were slightly better for smaller $N$, implying that smaller training datasets captured slightly better association between independent ($d$) and dependent variables ($t^{l}$ and $t^{c}$). However, the differences were slight, indicating that training data size did not substantially affect the correlation between independent and dependent variables. 

Hence, we can assume that a larger sample size does not imply a stronger correlation; in fact, in our case, a larger sample actually weakens the correlation by a slight degree. This goes to show that the homogeneity of the sample is more important than its size. How we collect the sample is also a key factor; if the data were sampled randomly, the correlation could have been lower. As we sequentially collected data during the experiment, the sample maintained a correlation comparable to larger datasets. Hence, we can conclude that smaller sequential datasets can be effective in predicting the execution times of robotics applications.

\subsection{Residuals}

Residuals (specifically, the difference between mean $t$ and mean $p$) help us determine the error between actual and predicted execution times. Residual values for both cloud and local execution for models trained on different dataset sizes are listed in Table \ref{tab:results}. With cloud computing, the residual  was around 0.03681 for $N=5$, which implies a 26$\%$ error rate (residual / actual $t^c$). Error rates decreased with increasing $N$, being about 13$\%$ for $N=10$, 4\% for $N=20$ \& $30$, and less than 1\% for all higher $N$ values. A less than 1\% error rate implies an average difference between actual and predicted execution times of less than 0.39 milliseconds. 

Meanwhile, error rate for predictions of local execution time for $N = 5$ featured a residual of 0.01572 and a resulting error rate of about 8\%, while all higher $N$ values had error rates of around 2\%. Due to the high correlation between local execution time and input data size, there was better convergence and prediction accuracy at lower $N$. The 2\% error rate implies an average difference of 3.5 milliseconds between actual and predicted execution times. 

One of the takeaways we want to highlight is that error percentages were high when $N=5$, but less than 2\% when $N >40$. This implies that complex algorithms trained on large datasets are not needed to accurately predict offloading decision-making. A simple lightweight regression algorithm fitted on the previous $N$ elements will be able to efficiently decide when to offload the application to the cloud. In this case, $N >40$ was sufficient to yield an acceptable level of error rate of less than 2\%. 

\subsection{Decision Making Accuracy}

To evaluate the accuracy of the algorithm, we compared the predicted action obtained from $p^{l}$ and $p^{c}$ (Eq. \ref{localexecutiontimere}) with the correct action obtained from $t^{l}$ and $t^{c}$. The accuracy was around 60\% for $N = 5$ and around 79\% for $N = 50$. Above $N = 50$, only marginal increases in accuracy were obtained. We also trained the model on the entire dataset ($N =1,000$) with 80:20 (training:testing) split, the result from which implies that the best possible accuracy for the linear regression will be around 83\% for the given dataset. 

We then further evaluated the proposed model by comparing it with other state-of-the-art machine learning models. Unlike in robotics, application offloading is well-studied in the context of mobile devices \cite{10.1007/s11036-012-0368-0,BHATTACHARYA201797}, which literature has predominantly used Long Short-Term Memory (LSTM)  for predicting execution time \cite{LU2020847}. Hence, we implemented a LSTM model based on \cite{MIAO2020925} (hidden layers = 4, batch size = 50) to predict application execution time from input data size. For training and testing, we used an 80:20 split of the entire dataset ($N =1,000$). We then compared the predicted actions obtained from LSTM predicted execution times with the actual actions obtained from $t^{l}$ and $t^{c}$. The LSTM model achieved a final accuracy of 75.68\%, whereas the linear regression achieved an accuracy of 83.32\%.  Hence, we can conclude that our predictive algorithm was comparable in accuracy to LSTM for $N =1,000$.

Observing the residual values and accuracy gives us reasonable confidence in the ability of the proposed algorithm to correctly predict execution times when trained on smaller datasets.

\subsection{Discussion}
In this paper, we propose a predictive algorithm for offloading decision-making, and further validate the algorithm using linear regression with a robot path planning application. 

Several exogenous factors can directly or indirectly affect the execution time of an application. It might not be practically feasible to capture all such variables and train a machine learning algorithm to incorporate them into highly-accurate predictions of execution time. These influencing factors are often periodical, such as when a new application is launched alongside currently-running applications, requiring all of the applications share computational resources and thus extending execution time until one or another application terminates. Another example is network failure, which will extend the cloud execution time until network connectivity is restored. These exogenous factors are incredibly difficult to predict, but by considering $N$ previous values we can capture some periodical factors and so predict execution times that are true to the current state of the system. Moreover, we observed a prediction accuracy of around 80\% with only 2\% difference between the actual and predicted values when we considered a training set of size $N = 40$. With such a small $N$ value, the predictive algorithm will be quicker to train, will output its predictions faster, and will require much data for training compared with traditional machine learning algorithms. 

One potential drawback of the proposed algorithm is that it only works effectively when both an independent variable and a significantly correlated dependent variable exist. Furthermore, the linear algorithm is usually sensitive to outliers, hence outliers must be appropriately addressed before fitting the data. Another drawback with a linear algorithm is that the data should have a linear relation. If the number of prediction variables is increased, issues with data linearity might occur. Finally, we also validated the proposed algorithm for subtasks that execute sequentially, but it remains to be seen how the algorithm performs for applications that execute in parallel. Perhaps parallel execution will need to be implemented in a distributed setup. Another drawback we want to highlight is that the accuracy of the prediction depends on the correlation between input data and predicted values; without a significant correlation, predicted values will be unreliable and can result in less-correct actions. 

Finally, we do not claim that these models outperform state-of-the-art machine learning models. Machine learning models trained on large datasets with a variety of features will outperform these predictive models. However, it is not always possible to have a pre-trained machine learning model for a variety of applications and robot hardware. Our proposed predictive model can be used as an alternative in situations where prior knowledge is not available and offloading decisions need be made on the fly as quickly as possible. A video on the paper is available for reference at: \\ {\textcolor{blue}{ \url{https://youtu.be/w3tniTpgjYY}}}.

\section{Conclusion} \label{sec:conclusion}

In this paper, we introduce a predictive algorithm to predict the execution time of an application for a given application data input size and then use that prediction for decision-making regarding application offloading. As the proposed algorithm starts learning after the application is initiated, minimizing training time is of highest priority; this is achieved by training the algorithm on a small number of previous data observations ($N$). We expect that this training approach will enable the capture of exogenous factors that are not directly incorporated in the model’s design. 

To validate the proposed predictive algorithm, we used linear regression and a Gazebo world. We experimented with varying $N$ values to evaluate performance, and we found that the algorithm had an acceptable error and prediction accuracy when $N >40$.

In future work, we will also consider additional cost parameters such as energy usage. We are also working on algorithms, such as ARIMA and SARIMA \cite{wei2006time}, that can help predict execution times for applications whose  variable relationships are non-linear.

\bibliographystyle{IEEEtran}
\bibliography{IEEEabrv,manoj}

\end{document}